\documentclass[letterpaper, 10 pt, conference]{ieeeconf}  %

\IEEEoverridecommandlockouts                              %

\usepackage{amssymb}            %
\usepackage{mathtools}          %
\usepackage{mathrsfs}           %

\usepackage{amsmath,amsfonts,bm}

\newcommand{\model}{DiffFP}

\def\eqref#1{equation~\ref{#1}}

\def\1{\bm{1}}

\DeclareMathAlphabet{\mathsfit}{\encodingdefault}{\sfdefault}{m}{sl}
\SetMathAlphabet{\mathsfit}{bold}{\encodingdefault}{\sfdefault}{bx}{n}

\DeclareMathOperator*{\argmax}{arg\,max}

\newtheorem{definition}{Definition}
\newtheorem{remark}{Remark}
\newcommand{\given}{\,|\,}

\usepackage[colorlinks=true, linkcolor=blue, urlcolor=blue]{hyperref}
\usepackage[acronym]{glossaries}
\glsdisablehyper
\newacronym{SGD}{\textsc{sgd}}{stochastic gradient descent}
\newacronym{MAP}{\textsc{map}}{maximum-a-posteriori}
\newacronym{MLE}{\textsc{mle}}{maximum likelihood estimation}
\newacronym{MNLL}{\textsc{mnll}}{mean negative log-likelihood}
\newacronym{NLL}{\textsc{nll}}{negative log-likelihood}
\newacronym{LL}{\textsc{ll}}{log-likelihood}
\newacronym{RMSE}{\textsc{rmse}}{root mean square error}
\newacronym{ECE}{\textsc{ece}}{expected calibration error}
\newacronym{SNR}{\textsc{snr}}{signal-to-noise ratio}
\newacronym{FID}{\textsc{fid}}{Fr\'echet Inception Distance}
\newacronym{BPD}{\textsc{bpd}}{bit per dimension}
\newacronym{NFE}{\textsc{nfe}}{neural function evaluations}

\newacronym{AE}{\textsc{ae}}{auto-encoder}
\newacronym{WAE}{\textsc{wae}}{Wasserstein Auto-encoder}
\newacronym{VAE}{\textsc{vae}}{Variational Auto-encoder}
\newacronym{BAE}{\textsc{bae}}{Bayesian Auto-encoder}
\newacronym{CDF}{\textsc{cdf}}{cumulative density function}
\newacronym{GAN}{\textsc{gan}}{Generative Adversarial Network}
\newacronym{DPGMM}{\textsc{dpgmm}}{Dirichlet process Gaussian mixture model}
\newacronym{GMM}{\textsc{gmm}}{Gaussian mixture model}

\newacronym{MC}{mc}{Monte Carlo}
\newacronym{SDE}{\textsc{sde}}{Stochastic Differential Equation}
\newacronym{CNF}{cnf}{Continuous Normaxlizing Flow}
\newacronym{ODE}{ode}{Ordinary Differential Equation}
\newacronym{MCMC}{\textsc{mcmc}}{Markov chain Monte Carlo}
\newacronym{HMC}{\textsc{hmc}}{Hamiltonian Monte Carlo}
\newacronym{MH}{mh}{Metropolis-Hastings}
\newacronym{NUTS}{nuts}{no-u-turn sampler}
\newacronym{SGHMC}{\textsc{sghmc}}{stochastic gradient Hamiltonian Monte Carlo}

\newacronym[longplural=deep Gaussian processes]{DGP}{\textsc{dgp}}{deep Gaussian process} %
\newacronym{GPLVM}{gplvm}{Gaussian process latent variable model}
\newacronym{DPMM}{dpmm}{Dirichlet Process Mixture Model}

\newacronym{VFE}{vfe}{variational free energy}

\newacronym[longplural=Gaussian Processes]{GP}{\textsc{gp}}{Gaussian Process}

\newacronym{VI}{\textsc{vi}}{variational inference}
\newacronym{SVI}{\textsc{svi}}{stochastic variational inference}

\newacronym{ELBO}{\textsc{elbo}}{evidence lower bound}
\newacronym{NELBO}{\textsc{nelbo}}{negative evidence lower bound}
\newacronym{ELL}{\textsc{ell}}{expected log likelihood}
\newacronym{KL}{\textsc{kl}}{Kullback-Leibler}
\newacronym{AUC}{auc}{area under the curve}

\newacronym{BNN}{\textsc{bnn}}{Bayesian neural network}
\newacronym{DNN}{\textsc{dnn}}{deep neural network}
\newacronym{CNN}{\textsc{cnn}}{convolutional neural network}
\newacronym{MLP}{\textsc{mlp}}{multilayer perceptron}
\newacronym{NN}{nn}{neural network}
\newacronym{RELU}{ReLU}{rectified linear unit}

\newacronym{NF}{nf}{normalizing flow}

\newacronym{RBF}{rbf}{radial basis function}
\newacronym{ARD}{ard}{automatic relevance determination}

\newacronym{RKHS}{rkhs}{reproducing kernel Hilbert space}

\newacronym{OT}{ot}{optimal transport}
\newacronym{WD}{wd}{Wasserstein distance}
\newacronym{SWD}{swd}{sliced-Wasserstein distance}
\newacronym{DSWD}{dswd}{distributional sliced-Wasserstein distance}
\newacronym{fsp}{FSP}{Fictitious Self Play}
\newacronym{marl}{MARL}{multi-agent reinforcement learning}
\newacronym{pomg}{POMG}{Partially Observable Markov Game}
\newacronym{ddpm}{DDPM}{Denoising Diffusion Probabilistic Models}
\newacronym{rl}{RL}{Reinforcement Learning}
\newacronym{sl}{SL}{Supervised Learning}
\newacronym{dp}{DP}{Diffusion Policy}
\newacronym{mpe}{MPE}{Multiple Particle Environment}
\newacronym{fp}{FP}{Fictitious Play}
\newacronym{mdp}{MDP}{Markov decision process}

\usepackage{graphicx}           %
\usepackage{subcaption}         %
\usepackage[space]{grffile}     %
\usepackage{url}                %
\usepackage{lipsum}             %
\usepackage{algorithm}
\usepackage{algorithmic}
\usepackage[dvipsnames]{xcolor}
\usepackage{multirow} 
\usepackage{cite}

\title{\LARGE \bf
DiffFP: Learning Behaviors from Scratch via
Diffusion-based Fictitious Play
}

\author{
    Akash Karthikeyan, Yash Vardhan Pant
    \thanks{The authors are with the Department of Electrical and Computer Engineering, University of Waterloo, Waterloo, Canada.
  \ttfamily{a9karthi@uwaterloo.ca, yash.pant@uwaterloo.ca}
}
}

\begin{document}

\maketitle
\thispagestyle{empty}
\pagestyle{empty}

\begin{abstract}
Self-play reinforcement learning has demonstrated significant success in learning complex strategic and interactive behaviors in competitive multi-agent games. However, achieving such behaviors in continuous decision spaces remains challenging. Ensuring adaptability and generalization in self-play settings is critical for achieving competitive performance in dynamic multi-agent environments.
These challenges often cause methods to converge slowly or fail to converge at all to a Nash equilibrium, making agents vulnerable to strategic exploitation by unseen opponents.
To address these challenges, we propose \textbf{\model}, a fictitious play (FP) framework that estimates the best response to unseen opponents while learning a robust and multimodal behavioral policy. Specifically, we approximate the best response using a diffusion policy that leverages generative modeling to learn adaptive and diverse strategies.
Through empirical evaluation, we demonstrate that the proposed FP framework converges towards $\bm\varepsilon$-Nash equilibria in continuous-space zero-sum games. We validate our method on complex multi-agent environments, including racing and multi-particle zero-sum games. Simulation results show that the learned policies are robust against diverse opponents and outperform baseline reinforcement learning policies. Our approach achieves up to \emph{3× faster convergence} and \emph{30× higher success rates} on average against RL-based baselines, demonstrating its robustness to opponent strategies and stability across training iterations~\footnote{Supplementary video: \href{https://aku02.github.io/projects/difffp/}{https://aku02.github.io/projects/difffp/}}.

\end{abstract}

\section{Introduction}
\label{sec:intro}
\gls{rl} enables the optimization of decision-making processes through interaction with simulators, where the environment is modeled as a \gls{mdp}. The objective is to learn an optimal policy that maximizes the expected cumulative reward over time. However, transitioning from the single-agent to the multi-agent setting introduces complex dynamics. The environment becomes non-stationary due to the presence of multiple learning agents, and interdependence among agents' strategies further complicates the learning process. Particularly in competitive games, achieving convergence to near-Nash equilibirum policies that are robust to exploitation is challenging, as exploration of the solution space in continuous domains becomes significantly harder~\cite{yang2024STLGame}.

Within game-theoretic frameworks, \gls{marl} has achieved remarkable success in domains such as Go, Atari, Chess, Shogi, and StarCraft~II~\cite{SilHub18General, schrittwieser2020mastering, moravvcik2017deepstack, alphastar19}, demonstrating the transformative impact of \gls{marl} in complex and dynamic environments. In these systems, agents learn strategies through policy optimization techniques, parameterized by deep neural networks. 

A common approach to learning robust policies in interactive competitive settings is self-play \cite{robinson_fp51}, where agents iteratively improve by playing against copies or past versions of themselves. \gls{fp}~\cite{fp51} is a game-theoretic algorithm that formalizes this. In \gls{fp}, each agent computes the \emph{best response} to the empirical average of its opponents' past actions. Under certain conditions, the sequence of strategies converges to a Nash equilibrium.
Recent advances have extended \gls{fp} to more complex settings,  such as dynamic environments, extensive-form games, and continuous-state Markov games~\cite{heinrich2015fictitious, nfsp-adapt23}. Fictitious Self-Play (FSP) provides an iterative procedure for converging to a Nash equilibrium, where each agent's strategy is a best response to the empirical distribution of opponents' strategies.

In practice, computing an exact best response in continuous and high-dimensional environments is challenging~\cite{yang2024STLGame}. \gls{rl}-based methods are typically employed to approximate the best responses; however, standard RL policies tend to be unimodal~\cite{yang2023policy}. This unimodality limits their ability to capture the inherent multimodality of complex tasks, often resulting in policies that overfit to recent opponent behaviors while neglecting previously encountered strategies~\cite{vinitsky2020robustreinforcementlearningusing}.

Recent work has demonstrated the power of diffusion models %
in capturing multimodal data distributions, with notable success in behavioral cloning, imitation learning, and planning~\cite{ho2020denoising, chi2023diffusion, janner2022planning}. However, most of the work focuses on learning from demonstration and their application in online RL, where agents learn from scratch, remains relatively under-explored. While some approaches leverage diffusion models in offline settings~\cite{wang2022diffusion, Ada_2024, kang2024efficient}, integrating them into the online RL framework is challenging due to the dynamic nature of value estimation as policies evolve and the diffusion objective cannot directly capture these changes~\cite{yang2023policy, psenka2023learning, ding2024diffusionbased}.

\noindent\textbf{Contributions of this work.} To overcome the challenges described above, we propose \model, a fictitious play framework that estimates best responses through a diffusion policy-based \gls{rl} approach. Our key contributions are as follows:
\begin{itemize}
    \item \model{} learns from scratch via iterative self-play, training a diffusion policy as a best response to an evolving average strategy.
    \item \model{} models the average strategy explicitly and update it online, enabling stable learning dynamics and reducing non-stationarity in multi-agent training.
    \item \model{} enables sample-efficient learning of multi-modal action distributions, leveraging the expressiveness of diffusion models to capture diverse behaviors in continuous action spaces.
\end{itemize}
Through extensive simulation-based studies on multi-agent games with continuous state and action spaces, we demonstrate that the learned policies converge to near Nash equilibria, as measured by low exploitability, and exhibit robustness to unseen opponents. In particular, we observe up to \emph{3× faster convergence} to near-optimal performance and \emph{30× higher success rates} on average compared to RL-based baselines.

\section{Related Work}
\label{sec:rlt}

\subsection{Adversarial Robustness}
Maximum Entropy \gls{rl}~\cite{haarnoja2018soft} encourages robustness by maximizing expected reward while maintaining high entropy in the policy distribution. Though this has proven effective in standard continuous control tasks~\cite{todorov2012mujoco}, it struggles in dynamic multi-agent settings where adversarial behaviors lead to non-stationary learning targets. To address these challenges, population-based training frameworks inspired by empirical game theory~\cite{wellman2006} have been proposed. These maintain a pool of evolving strategies and compute best responses, as in Policy-Space Response Oracles (PSRO)~\cite{lanctot}. 

In continuous control tasks,~\cite{Bansal2017EmergentCV} propose opponent sampling, where agents train against past versions of opponents using various sampling strategies similar to \gls{fp}, but do not provide exploitability studies.
 Similarly, Adversarially Robust Control (ARC)~\cite{Kuutti} tackles robustness in semi-competitive autonomous driving settings by training agents against an ensemble of adversaries, while regularizing towards imitation-learned behaviors to prevent overfitting. 
 
More recent approaches, such as population-based diversity shaping~\cite{gleave23}, explicitly target diversity and robustness in continuous control via population-based training. However, these methods often incur additional computational cost and remain sensitive to adversary tuning, with diminishing returns observed as population size increases~\cite{vinitsky2020robustreinforcementlearningusing}. In general, when computing best responses under a finite budget, approximate best-response operators must be truncated at each iteration, resulting in under-trained policies populating the population. Furthermore, relearning policies from scratch at each iteration becomes intractable and hinders scalability to larger action spaces~\cite{liu2022neupl}

A range of multi-agent RL approaches aim to produce policies that are less exploitable than those obtained via standard self-play. One prominent example is Counterfactual Regret Minimization (CFR)~\cite{zinkevich2007}, which has achieved superhuman performance in games like poker~\cite{brown2018}. However, extending CFR to continuous-action games with high-dimensional state and action spaces remains an open challenge. In contrast, self-play continues to be a dominant paradigm due to its scalability and empirical success across a wide range of complex domains~\cite{silver2017mastering, alphastar19}.

\subsection{Game-Theoretic Learning}
Self-play can, in principle, converge to a Nash equilibrium if reinforcement learning produces a best response at each iteration~\cite{heinrich2015fictitious}. In practice, convergence is often elusive due to challenges such as \emph{non-transitivity}, as in rock-paper-scissors, where learning dynamics may enter persistent cycles~\cite{balduzzi_openended_2019}. Even in transitive games, progress can stall because of local minima or the limited expressivity of the policy architecture.

Neural Fictitious Self-Play~\cite{nfsp16} provides a general framework for multi-agent systems through end-to-end deep reinforcement learning and has recently been extended to continuous-action settings, where best responses are approximated using deep reinforcement learning~\cite{nfsp-adapt23}. Such extensions are particularly relevant in domains such as robotics and autonomous driving, which involve continuous action spaces, partial observability, and imperfect information.

Inspired by \gls{fsp} and its neural parametrization, we extend \gls{fp} to high-dimensional continuous-control tasks and learn robust best-response policies \emph{from scratch} via diffusion-based policy representations. These representations are capable of capturing multimodal strategies, avoiding cyclic forgetting, and improving sample efficiency. Our formulation accommodates both symmetric and asymmetric games, enabling scalable learning with improved sample efficiency and generalization.

\section{Preliminaries and Problem Statement}
\label{sec:prelims}
We consider a \gls{pomg}~\cite{posg04} defined by the tuple 
$(\mathcal{I}, \mathcal{S}, b^0, \{\mathcal{A}_i\}, \{\mathcal{O}_i\}, \mathcal{T}, \Omega_i, \{R_i\}),$
where:
\begin{itemize}
    \item $\mathcal{I} = \{1, \cdots, n\}$ is the finite set of agents;
    \item $\mathcal{S}_i \subseteq \mathbb{R}^{d_s}$ is the continuous state space for agent $i$, and $\mathcal{S} = \prod_{i \in \mathcal{I}} \mathcal{S}_i$ is the joint state space, with $\overrightarrow{s} = (s^1, \cdots, s^n) \in \mathcal{S}$ denoting the joint state;
    \item $b^0 \in \mathcal{P}(\mathcal{S})$ denotes the initial state distribution;
    \item $\mathcal{A}_i \subseteq \mathbb{R}^{d_a}$ is the continuous action space for agent $i$, and $\mathcal{A} = \prod_{i \in \mathcal{I}} \mathcal{A}_i$ is the joint action space, with $\overrightarrow{a} = (a^1, \cdots, a^n) \in \mathcal{A}$;
\item $\mathcal{O}_i$ is the observation space for agent $i$, with observation function $\Omega_i : \mathcal{S} \to \mathcal{P}(\mathcal{O}_i)$ mapping the global state to the agent's local observation (e.g., an occupancy grid);
    \item $\mathcal{T}: \mathcal{S} \times \mathcal{A} \to \mathcal{P}(\mathcal{S})$ is the (joint) transition function; 
    \item $R_i: \mathcal{S} \times \mathcal{A} \to [r_{\min}, r_{\max}] \subset \mathbb{R}$ is the bounded reward function for agent $i$, where
    \( r_{\min} \leq R_i(\overrightarrow{s}, \overrightarrow{a}) \leq r_{\max} \)
    for all \( (\overrightarrow{s}, \overrightarrow{a}) \in \mathcal{S} \times \mathcal{A} \).
\end{itemize}

\textbf{A (stochastic) policy} for agent \(i\) is defined as a mapping 
\(\pi^i: \mathcal{O}_i \to \mathcal{P}(\mathcal{A}_i)\), 
where \(\mathcal{P}(\mathcal{A}_i)\) denotes the probability measures over 
the action space \(\mathcal{A}_i\). 
The joint strategy of all agents is denoted as 
\(\boldsymbol{\pi} := (\pi^1, \ldots, \pi^n) = (\pi^i, \boldsymbol{\pi}^{-i})\), 
where \(\boldsymbol{\pi}^{-i}\) represents the strategy of all agents except agent \(i\).

\begin{remark}[Two-player zero-sum game]
We consider a two-player zero-sum game, where the sum of agents' rewards satisfies $(\sum_i R_i = 0)$. Without loss of generality, this formulation extends naturally to multi-agent settings. The joint strategy is denoted as \(\boldsymbol{\pi} = (\pi^{\text{ego}}, \pi^{\text{opp}})\), where \(\pi^{\text{ego}}\) and \(\pi^{\text{opp}}\) represent the behavioral strategies of the ego and opponent agents, respectively.
\end{remark}

\begin{definition}[Best Response]
Given the opponent's strategy \(\boldsymbol{\pi}^{-i}\), the best-response strategy \(\pi^i_{\mathrm{BR}}\) for agent \(i\) is defined as
\begin{equation}
\pi^i_{\mathrm{BR}} \in \argmax_{\pi^i \in \Pi^i} J_i(\pi^i, \boldsymbol{\pi}^{-i}),
\end{equation}
where \(\Pi^i\) denotes the set of admissible policies for agent \(i\), and 
\(J_i: \Pi^1 \times \dots \times \Pi^n \to \mathbb{R}\) denotes the expected cumulative reward for agent \(i\).
\end{definition}
\vspace{0.1cm}

In \emph{zero-sum games}~\cite{vonneumann1944theory, littman1994markov}, the reward functions satisfy \( R^{\text{ego}} = -R^{\text{opp}} \), resulting in a minimax formulation where one agent's gain is the other's loss. The goal is to compute joint strategies that are robust against adversarial opponents.
\begin{definition}[$\bm\varepsilon$-Nash Equilibrium] 
We define a joint strategy profile \(\boldsymbol{\pi}^* = (\pi^{\text{ego},*}, \pi^{\text{opp},*})\) to be an \(\bm\varepsilon\)-Nash equilibrium if, for all \( i \in \{\text{ego}, \text{opp}\} \), the following holds:
\begin{equation}
J_i\big(\pi^{i,*}, \boldsymbol{\pi}^{-i,*}\big) \ge J_i\big(\pi^i, \boldsymbol{\pi}^{-i,*}\big) - \bm\varepsilon, 
\quad \forall\, \pi^i \in \Pi^i.
\end{equation}
\end{definition}
\vspace{0.1cm}
\noindent Intuitively, this means that no agent can gain more than \(\bm\varepsilon\) by unilaterally deviating from their current strategy.
\noindent The deviation from an exact equilibrium can be quantified using the notion of \emph{exploitability}, which measures how much an agent's return could be improved by switching to a best-response policy.
\begin{definition}[Exploitability]\label{def:exp} 
We define the exploitability of a joint policy profile \(\boldsymbol{\pi} = (\pi^i, \pi^{-i})\) for agent \(i\) as:
\begin{equation}
\label{eq:exp}
\bm\varepsilon_i = J_i\big(\pi_{\mathrm{BR}}^i, \boldsymbol{\pi}^{-i}\big) - J_i\big(\pi^i, \boldsymbol{\pi}^{-i}\big),
\end{equation}
where \(\pi_{\mathrm{BR}}^i\) is a best response to \(\pi^{-i}\). The total exploitability is given by \(\bm\varepsilon = \sum_{i \in \mathcal{I}} \bm\varepsilon_i\).
\end{definition}

\subsection{Problem Statement}
\label{sec:minimax_obj}
We consider a two-agent zero-sum \gls{pomg} where agents interact in a continuous state and action space. Each agent \( i \in \{\text{ego}, \text{opp}\} \) seeks to maximize its expected return $(J_i(\pi^i, \boldsymbol{\pi}^{-i}))$ over a horizon \( H \), under the assumption that its opponent behaves adversarially. This leads to a minimax optimization problem defined over joint policies.
\begin{equation*}
\label{eq:expected_return}
J_i(\pi^i, \boldsymbol{\pi}^{-i}) = \mathbb{E}_{\substack{
\overrightarrow{a_h} \sim \boldsymbol{\pi}(\cdot \mid \overrightarrow{o_h}), \; \\
\overrightarrow{{s}_{h+1}} \sim \mathcal{T}(\cdot \mid \overrightarrow{{s}_h}, \overrightarrow{{a}_h})}}
\left[ \sum_{t=0}^{H} \gamma^t\, R_i(\overrightarrow{{s}_h}, \overrightarrow{{a}_h}) \right],
\end{equation*}
The goal of each agent is to solve the following minimax optimization:
\begin{equation}
\label{eq:minimax_obj}
\pi^{*, i} = \arg\max_{\pi^i \in \Pi^i} \; \min_{\pi^{-i} \in \Pi^{-i}} \; J_i(\pi^i, \pi^{-i}),
\end{equation}
\noindent Without loss of generality, this formulation extends naturally to multi-agent settings involving more than two agents or teams. In such cases, each agent maximizes its own return while accounting for the joint policies of all other agents, which together act adversarially in the zero-sum context.

\section{\model: Method}
\label{sec:method}
\noindent\textbf{Notations.} 
We use subscript $h$ to denote timesteps in the \gls{pomg}, and $t \in \{1, \ldots, T\}$ for diffusion timesteps. Agent index is indicated by superscript $i$. The action taken by agent $i$ at environment timestep $h$ and diffusion step $t$ is denoted as $a^{i}_{h,t}$. Under fictitious play, we denote the policy of agent $i$ at iteration $k \in \{0, \ldots, K\}$ as $\pi^i_k$. We use $\mathcal{U}\{1, \dotsc, T\}$ to denote an uniform distribution over the set $\{1, \dotsc, T\}$, and $\mathcal{N}$ to denote a Gaussian distribution. The symbol $\,\sim\,$ denotes sampling from a distribution; for example, $\epsilon \sim \mathcal{N}(0, I)$. \textit{For brevity, we drop the subscript $h$ or $t$ in equations where all terms refer to the same \gls{pomg} timestep or diffusion step.}

To solve the minmax problem in continuous zero-sum games described in Section~\ref{sec:minimax_obj}, we adopt a \gls{fp} style learning framework. Specifically, we integrate diffusion models as expressive policy classes to represent the agents' stochastic strategies in high-dimensional, continuous action spaces. We train a diffusion-based policy at each iteration to approximate the best response against a fixed opponent strategy distribution. 

\glsreset{fp}
\subsection{\gls{fp}} %
\label{sec:fsp_method}
\glsdesc{fp}~\cite{fp51} is an iterative algorithm in which agents update their strategies based on the empirical distribution of their opponents' past strategies. At each iteration $k$, each agent $i$ computes a best response $\pi_{\mathrm{BR}}^i$ to the current average strategy $\boldsymbol{\pi}^{-i}_k$ of its opponents and updates its strategy
\begin{equation}
\label{eqn:fsp_avg}
\nonumber
\pi^i_{k+1} = \frac{k}{k+1}\,\pi^i_k + \frac{1}{k+1}\,\pi_{\mathrm{BR}}^i.    
\end{equation}

One of the primary challenges in fictitious play lies in computing the best response policy, particularly in environments with continuous action spaces. To address this, we employ a generalized weakened fictitious play framework~\cite{leslie2006generalised}, which allows for approximate best responses and small perturbations. Formally, a sequence of mixed strategies ${\pi_k}$ follows generalized weakened fictitious play if, for each agent $i$,
$$
\pi_{k+1}^i \in (1-\alpha_{k+1})\,\pi_k^i + \alpha_{k+1}\Big(\textsc{BR}_{\epsilon_k}(\boldsymbol{\pi}_k^{-i}) + M_{k+1}^i\Big),
$$
where the step size $\alpha_{k}\rightarrow 0$ and $\epsilon_k\rightarrow 0$, when $\sum_{k=1}^{\infty}\alpha_{k}=\infty$, and $M_{k}^i$ is a perturbation term satisfying, for any constant $J>0$,
$$
\lim_{k\to\infty} \sup_{j}\left\{\left\|\sum_{l=k}^{j-1} \alpha_{l+1} M_{l+1}^i \right\| \textit{s.t.} \sum_{l=k}^{j-1} \alpha_{l+1}\leq J \right\} = 0, \forall J \geq 0.
$$

Algorithm~\ref{algorithm:fp_fsp} summarizes the \glsdesc{fp} procedure in our framework.
By iteratively refining the strategies through \gls{fp} where RL is employed to approximate best responses we ensure that the overall exploitability decreases over time, thereby converging to an $\bm\varepsilon$-Nash equilibrium. 
\begin{algorithm}[b]
\caption{Fictitious Self-Play}
\label{algorithm:fp_fsp}
\begin{algorithmic}[1]
    \STATE \textbf{Initialize} $\pi^i_0$ randomly for each agent $i \in \{\text{ego}, \text{opp}\}$.
    \FOR{$k = 0, \ldots, K-1$}
        \FOR{each agent $i \in \{\text{ego}, \text{opp}\}$}
            \STATE Compute best response: $\pi_{\mathrm{BR}}^i \gets \textsc{BR}(\boldsymbol{\pi}_k^{-i})$.
            \STATE {\color{blue}{\texttt{\# Compute $\mathrm{BR}$ through Alg.~\ref{algo:br}}}}
            \STATE Update strategy: $\pi^i_{k+1} \gets \frac{k}{k+1}\,\pi^i_k + \frac{1}{k+1}\,\pi_{\mathrm{BR}}^i$.
        \ENDFOR
    \ENDFOR
\end{algorithmic}
\end{algorithm}
\subsection{Approximating Best Response via \gls{rl}}
\begin{algorithm}[tb]
    \caption{\model{} as Best Response for Agent \(i\)}
    \label{algo:br}
    \begin{algorithmic}[1]
        \STATE \textbf{Initialize} actors $\{\pi_\theta^i, \pi_{\theta'}^i\}$ and critics $\{Q_\psi^i, Q_{\psi'}^i\}$ for all $i \in \mathcal{I}$ with targets, buffer $\mathcal{D} \gets \emptyset$, Set noise schedule $\{\alpha_t\}_{t=1}^{T}$ with $\bar{\alpha}_t = \prod_{j=1}^t \alpha_j$, step size $\eta$, train steps $M$
        
        \FOR{iteration $m = 1$ to $M$}
            \STATE {\color{blue}{\texttt{\# Experience Collection}}}
            \FOR{timestep $h = 0$ to $H$}
                \STATE {\color{blue}{\texttt{\# Sample actions following eq.~\ref{eq:reverse_sampling}}}}
                \STATE Sample actions: $a_h^i \sim \pi^i(\cdot | o_h^i)$, $a_h^{-i} \sim \pi^{-i}(\cdot | o_h^{-i})$
                \STATE Construct joint action: $\overrightarrow{{a}_h} = (a_h^i, a_h^{-i})$
                \STATE Transition: $\overrightarrow{{s}_{h+1}} \sim \mathcal{T}(\cdot \mid \overrightarrow{{s}_h}, \overrightarrow{{a}_h})$
                \STATE Reward: $r^i_h = R_i(\overrightarrow{{s}_h}, \overrightarrow{{a}_h})$
                \STATE Store $(s_h^i, a_h^i, s_{h+1}^i, r^i_h)$ in buffer $\mathcal{D}$
            \STATE {\color{blue}{\texttt{\# Train Q-function}}}~\cite{hasselt2010double}
            \STATE Update parameters $\psi$ following equation~\ref{eq:q_update_compact}
            \STATE {\color{blue}{\texttt{\# Train Diffusion Policy}}}
            \STATE Update parameters $\theta$ using the loss in~\eqref{eq:diff_loss}
            \STATE {\color{blue}{\texttt{\# Q-guided Action Refinement}}}
            \FOR{$n = 1$ to $N_{\text{action-ascent}}$}
                \STATE $\tilde{a}_h^{n+1} \gets a_h^{n} + \eta \nabla_a Q_{\psi}(s_h, a)\big|_{a = a_h^{n}}$
                \STATE Add refined action to buffer: $\mathcal{D} \gets \mathcal{D} \cup \{(s_h, \tilde{a}_h^{n})\}$
            \ENDFOR            
            \ENDFOR
        \ENDFOR
    \end{algorithmic}
\end{algorithm}

Finding an exact best response is often computationally intractable, especially in large or continuous spaces. Prior works~\cite{heinrich2015fictitious, leslie2006generalised} have shown that instead of full-width backups, one can approximate the best response using \gls{rl}, and similarly approximate the average strategy using supervised learning (SL)~\cite{heinrich2015fictitious}.
\noindent\subsubsection{\textbf{Policy Representation}}
To model complex and multimodal team dynamics, we propose leveraging a diffusion policy~\cite{yang2023policy} to approximate the best response. While centralized modeling of joint team behavior offers computational benefits, it often falls short in capturing the inherent multi-modality of optimal responses, particularly in competitive or dynamic environments. Diffusion models address this limitation by learning diverse behavior distributions, making them especially effective at representing the rich variability in agent interactions. At each iteration $k$ of \gls{fp}, we compute the best response against a fixed but unknown average opponent policy, denoted by $\pi^{-i}_k$. For brevity, when discussing diffusion-based best responses, we omit the agent index and \gls{fp} index; the subscript $t$ in the following denotes the diffusion timestep.

\noindent\textbf{\gls{ddpm}}~\cite{ho2020denoising, nichol2021improved} form the backbone of this approach. \gls{ddpm} defines a forward process that gradually perturbs the data, here an action ${a}_0 \sim q({a}_0)$ drawn from the replay buffer is transformed into a tractable Gaussian prior $q({a}_T)$ over $T$ steps by iteratively applying Gaussian noise: 
\begin{equation}
\label{eq:diffusion_fw_slow}
q({a}_t | {a}_{t-1}) := \mathcal{N}({a}_t; \sqrt{1-\beta_t} {a}_{t-1}, \beta_t {I}), \forall t \in \{1,\dotsc,T\}
\end{equation}
where $\beta_t \in (0, 1)$ is the variance schedule. Through Markov chain property of the forward process, the distribution of any intermediate noisy action ${a}_t$ can be expressed in closed form conditioned on the original action ${a}_0$:
\begin{equation}
\label{eq:diffusion_fw}
q({a}_t | {a}_0) := \mathcal{N}({a}_t; \sqrt{\bar{\alpha}_t} {a}_0, (1-\bar{\alpha}_t) {I}), \forall t \in \{1,\dotsc,T\}
\end{equation}
with $\alpha_t = 1 - \beta_t$ and $\bar{\alpha}_t = \prod_{s=1}^t \alpha_s$. The forward diffusion process (Equations~\ref{eq:diffusion_fw_slow} and~\ref{eq:diffusion_fw}) progressively transforms clean actions into an isotropic Gaussian distribution as $T \to \infty$. The learned reverse process then inverts this trajectory to generate multi-modal actions from the policy.

The reverse process, corresponds to our policy $\pi_{\theta}(a_h|o_h) = p_{\theta}(a_{h,{0:N}}|o_h)$ begins from a sample ${a}_T \sim p({a}_T) = \mathcal{N}({0}, {I})$ drawn from the prior and conditionally denoises via a sequence of parametrized transitions: $p_\theta ({a}_{h,t-1} | {a}_{h,t}, o_h) := \mathcal{N}({a}_{h,t-1}; \mu_\theta({a}_{h,t}, o_h, t), \sigma_t^2 I)$ 
where $\theta$ denotes the parameters of a neural network trained to approximate the true reverse dynamics $q({a}_{h,t-1}|{a}_{h,t}, {a}_{h,0})$. 
The training objective is to maximize the Variational Lower Bound (VLB):
$L_{\mathrm{VLB}} := \mathbb{E}_{q({a}_{0:T})}\left[\log \frac{p_\theta({a}_{0:T})}{q({a}_{1:T} | {a}_0)}\right]$.
In practice, this objective simplifies to a conditional denoising score matching loss:
\begin{equation}
\label{eq:diff_loss}
\mathbb{E}_{t \sim \mathcal{U}[1,T], {a}_0, {\epsilon}}\left[||{\epsilon} - {\epsilon}_\theta(\sqrt{\bar{\alpha}_t} {a}_{0,h} + \sqrt{1 - \bar{\alpha}_t} {\epsilon}, t, o_h)||^2 \right]
\end{equation}
where $\epsilon_{\theta}(a_{h,t}, o_h, t) = \frac{\sqrt{1 - \bar{\alpha}_t}}{\beta_t} \left(a_{h,t} - \sqrt{\alpha_t}\,\mu_{\theta}(a_{h,t}, o_h, t)\right)$ is a neural network trained to predict the noise added during the forward process.
\begin{remark}[CTDE for Multi-Agent Teams]
Modeling multi-agent teams under the centralized training with decentralized execution (CTDE) paradigm presents significant challenges~\cite{wang2023more}. Within the same team, agents must learn complementary roles and diverse behaviors, resulting in inherently multi-modal action distributions. This coordination complexity makes it difficult for conventional methods to learn effective policies, while diffusion-based approaches are better suited to capture such diversity.
\end{remark}

\noindent\subsubsection{\textbf{Policy Learning}} Unlike a unimodal Gaussian policy, \gls{dp} does not maintain an explicit likelihood function. To learn multimodal policies for each agent $i$, we adopt the framework from~\cite{yang2023policy}, which combines a behavior cloning loss (\eqref{eq:diff_loss}) with double Q-learning~\cite{hasselt2010double} to guide policy improvement. 
\begin{equation}
\label{eq:q_update_compact}
\begin{aligned}
&\mathbb{E}_{(s_h^i, a_h^i, s^i_{h+1}) \sim \mathcal{D},\, a_{h+1,0}^i \sim \pi^i} \big[ \big(\big( 
r^i_h +  \\ & \gamma \min_{n=1,2} Q^i_{\psi_n'}(s^i_{h+1}, a^i_{h+1,0})\big)
- Q^i_{\psi_n}(s_h^i, a_h^i) \big)^2 \big]
\end{aligned}
\end{equation}
Policy improvement is achieved through action gradients that refine actions sampled from the replay buffer (line 17 of Alg~\ref{algo:br}). Additionally, we adopt reward-weighing~\cite{peters2007reinforcement} to prioritize higher-return samples and improve sample efficiency.

\noindent\subsubsection{\textbf{Sampling}} We iteratively recover the action following the reverse diffusion process. The actor is parameterized as a conditional diffusion model, defined as $\pi_{\theta}(a_h|o_h) := p_{\theta}(a_{h,0:T}|a_{h,T}, o_h)$. We retain only the denoised sample \( a_{h,0} \), which serves as the sampled action.
\begin{align}
\label{eq:reverse_sampling}
a_{t-1} \given a_t &= \frac{a_t}{\sqrt{\alpha_t}} - \frac{\beta_t}{\sqrt{\alpha_t(1 - \bar{\alpha}_t)}} \epsilon_\theta(a_{t}, o_h, t) + \sqrt{\beta_t} \epsilon, \\
\epsilon &\sim \mathcal{N}({0}, I), \quad \text{for }t=T,\dotsc,1. \nonumber
\end{align}
\noindent\textbf{Implementation.} We implement both the diffusion actor and critic networks using multilayer perceptrons (MLPs). The actor performs iterative denoising to generate actions, while the critic estimates the corresponding target value functions~\cite{hasselt2010double}. The overall best-response computation for an agent during each \gls{fp} iteration is summarized in Algorithm~\ref{algo:br}.
To compute the best response, we fix the opponent's average policy, which is stochastic and not explicitly known to the ego agent. Rather than approximating this average via supervised learning~\cite{heinrich2015fictitious}, we follow the exact averaging rule from~\eqref{eqn:fsp_avg} by sampling from the pool of historical best-response policies, weighted by their update coefficients as seen in line 5 of Algorithm~\ref{algorithm:fp_fsp}.

\section{Simulation Studies}
\label{sec:results}
We aim to answer the following questions through extensive simulation-based experiments:
\begin{itemize}
    \item \textbf{Convergence.} Can \model{} learn an approximately unexploitable policy profile? Specifically, in continuous and dynamic environments, can it converge to a joint $\bm\varepsilon$-Nash equilibrium?
    
    \item \textbf{Efficiency.} What is the benefit of using a \gls{dp}-based approach for modeling continuous-action zero-sum games, in terms of stability and sample efficiency?
    
    \item \textbf{Robustness.} Can our method learn diverse yet robust strategies that generalize to unseen opponents?
\end{itemize}
We evaluate \model{} on two benchmark simulation environments: one featuring symmetric agents with dense rewards, and the other characterized by asymmetric roles and sparse rewards. However, both settings are imperfect information games. All experiments are implemented in Python~3.8 and executed on a 12-core CPU with an RTX A6000 GPU~\footnote{Supplementary video: \href{https://aku02.github.io/projects/difffp/}{https://aku02.github.io/projects/difffp/}}.
\begin{figure*}[tb]
    \centering
    \includegraphics[width=\linewidth]{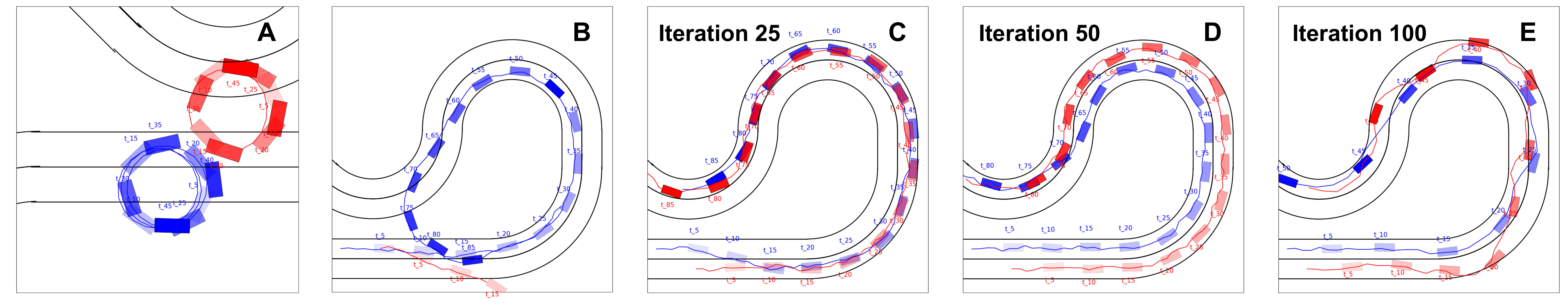}
    \caption{\textbf{(A–B) Failure Modes of Baselines.} Baseline agents exhibit suboptimal behaviors such as stalling or inefficient path planning. \textbf{(C–E) Training Progression of \model{}.} Agent trajectories sampled from successive FP iterations illustrate the emergence of strategic behaviors. Over time, agents learn to navigate more efficiently, reducing their time-to-goal from 80 to 45 steps under identical initial conditions.}
    \label{fig:racing_env}
\end{figure*}
\subsection{Baselines}
\label{sec:baselines}
We evaluate \model{} against several reinforcement learning (\gls{rl}) baselines, adapting them as (approximate) best-response within the \gls{fp} framework (Algorithm~\ref{algorithm:fp_fsp}).
\begin{itemize}
    \item \textbf{QSM}~\cite{psenka2023learning}: A generative (score-based) actor model trained by aligning the score function with the state-action gradient field, enabling policy learning via denoising objectives.

    \item \textbf{SAC}~\cite{haarnoja2018soft}: A maximum entropy \gls{rl} algorithm that estimates a stochastic policy by maximizing a trade-off between expected return and entropy.
    
    \item \textbf{TD3}~\cite{fujimoto2018td3}: A deterministic actor-critic method that addresses overestimation bias in Q-learning through clipped double Q-learning and target policy smoothing.
\end{itemize}
\noindent\textbf{Metrics.} We evaluate best responses using the total exploitability of the policy, as defined in Equation~\ref{eq:exp}. In practice, we use normalized episodic cumulative reward as the payoff measure. Additionally, we report normalized cumulative rewards for the racing task.
\subsection{Racing Environment.}
\noindent\subsubsection{\textbf{Setup}} We consider a multi-agent racing scenario in which agents must learn strategic behaviors such as overtaking, defending, and racing along a track. The interaction is modeled as a general-sum game~\cite{nash_general_sum}, where each agent is rewarded for minimizing the arc-length gap to its opponent while being constrained by track limits and collision avoidance. This yields a dense reward function that provides continuous guidance throughout the episode. The introduction of the collision-avoidance penalty augments each agent’s cost to encode safety constraints, yet the fictitious play fixed point remains equivalent to the first-order optimality conditions of a Nash equilibrium in the original (unpenalized) game~\cite{spica2018real, general_nash}.
We evaluate robustness by testing trained policies against unseen opponents across diverse track layouts. 
\noindent\subsubsection{\textbf{Results}} 
\begin{figure*}[tb]
    \centering
    \includegraphics[width=\linewidth]{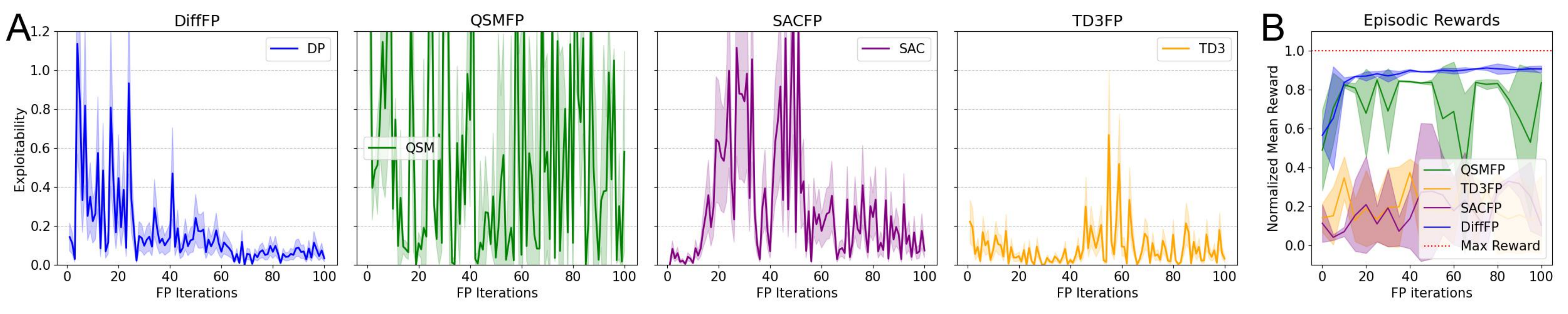}
    \caption{\textbf{A. Exploitability on the Racing Task.} Mean and standard deviation of exploitability over 10 episodes per \gls{fp} iteration. 
\textbf{B. Normalized Episodic Rewards.} This metric reflects training performance and stability.}
    \label{fig:racing_exp}
\end{figure*}

We report exploitability convergence in Figure~\ref{fig:racing_exp}. \model{} consistently outperforms all baselines, exhibiting both faster and more stable convergence toward low exploitability (see Definition~\ref{def:exp}).
\paragraph{\textbf{Limitations of Off-Policy RL}} Although TD3~\cite{fujimoto2018td3} and SAC~\cite{haarnoja2018soft} appear to converge in terms of exploitability metrics, qualitative evaluations reveal significant behavioral failures. As illustrated in Fig.~\ref{fig:racing_env}\textbf{A} and \textbf{B}, TD3 agents often fall into loop-like trajectories, circling without track progress, while SAC agents frequently violate track boundaries, failing to adhere to racing constraints. These behaviors highlight fundamental limitations of standard off-policy RL methods: they are typically Q-value mode-seeking and lack the expressiveness required to represent diverse and strategic behaviors~\cite{wang2022diffusion}. Moreover, despite large adversarial populations, they tend to overfit to specific opponents, resulting in brittle policies that remain exploitable. 
While increasing the adversary population can help introduce more varied dynamics, it incurs diminishing returns due to reduced environment interaction per policy~\cite{vinitsky2020robustreinforcementlearningusing}.
\noindent\paragraph{\textbf{Instability of Q-Function}} Generative approaches based on diffusion models~\cite{wang2022diffusion} offer greater expressiveness and naturally support multi-modal policy distributions, promoting behavioral diversity~\cite{yang2023policy}. However, QSM suffers from high instability due to its strong reliance on accurate value function estimates. This is particularly problematic in multi-agent settings, where the opponent's actions induce non-stationary dynamics, making Q-value estimation noisy and unreliable. As a result, exploitability convergence plots tend to be noisy and non-convergent.
\noindent\paragraph{\textbf{Track Progress}} Figure~\ref{fig:racing_exp} reports the normalized track progress reward. \model{} consistently achieves near-optimal performance, highlighting its robustness and stability in learning strategic racing behaviors.
\noindent\paragraph{\textbf{Training Progression}} 
Figure~\ref{fig:racing_env}(\textbf{C–E}) illustrates agent trajectories across \gls{fp} iterations. Initially (\textbf{C}), the blue agent trails behind. Over time (\textbf{D}), it learns to race competitively, overtaking on the inside and forcing the red agent to brake. Near convergence (\textbf{E}), both agents display aggressive and reactive behaviors:attacking, defending, and driving each other toward track boundaries, while completing the sector faster than in earlier iterations.

\begin{table}[tb]
    \centering
    \renewcommand{\arraystretch}{1.0}
    \setlength{\tabcolsep}{6pt} %
    \begin{tabular}{lcccc}
        \hline
        \textbf{Attacking} & \textbf{Gap (m)} & \textbf{Successes}& \textbf{Crashes} & \textbf{Mean Reward} \\
        \hline
        QSMFP      & 8.26  & 3 / 20  & 4 / 20 & 0.80 \\
        \model{} & \textbf{17.70} & \textbf{12 / 20} & \textbf{1 / 20} & \textbf{0.92} \\
        \hline
    \end{tabular}
    \caption{\textbf{Head-to-Head Trials.} Gap denotes the mean relative distance gained by the attacker. Successes refer to the number of episodes where the attacker successfully closed the initial gap or overtook the target. Attaking agent is intialized behind the race leader.}
    \label{tab:algo_comparison_single}
\vspace{-3mm}
\end{table}

\begin{table}[tb]
    \centering
    \renewcommand{\arraystretch}{1.2}
    \begin{tabular}{lrr}
        \hline
        \textbf{Opponent} & \textbf{Crashes} & \textbf{Mean Reward} \\
        \hline
        QSMFP & 11 / 20 & 0.556 \\
        \model{}  & \textbf{4 / 20}  & \textbf{0.756}  \\
        \hline
    \end{tabular}
    \caption{\textbf{Robustness to Unseen Opponents.} Evaluation of crash rates (lower is better) and mean rewards (higher is better) when the agent is tested against up to five previously unseen adversaries.}
    \label{tab:robustness_opponents}
\end{table}

\noindent\paragraph{\textbf{Robustness to Unseen Opponents}} Table~\ref{tab:algo_comparison_single} reports head-to-head results, where \model{} demonstrates stronger robustness by achieving more successful overtakes and fewer collisions with the race leader compared to the QSM baseline, which exhibits higher crash rates. In scenarios with 3–5 additional agents (Table~\ref{tab:robustness_opponents}), \model{} reliably avoids collisions and track violations through timely lane changes or braking, while baselines often exhibit unsafe behaviors.

\begin{figure}[tb]
    \centering
    \includegraphics[width=0.99\linewidth]{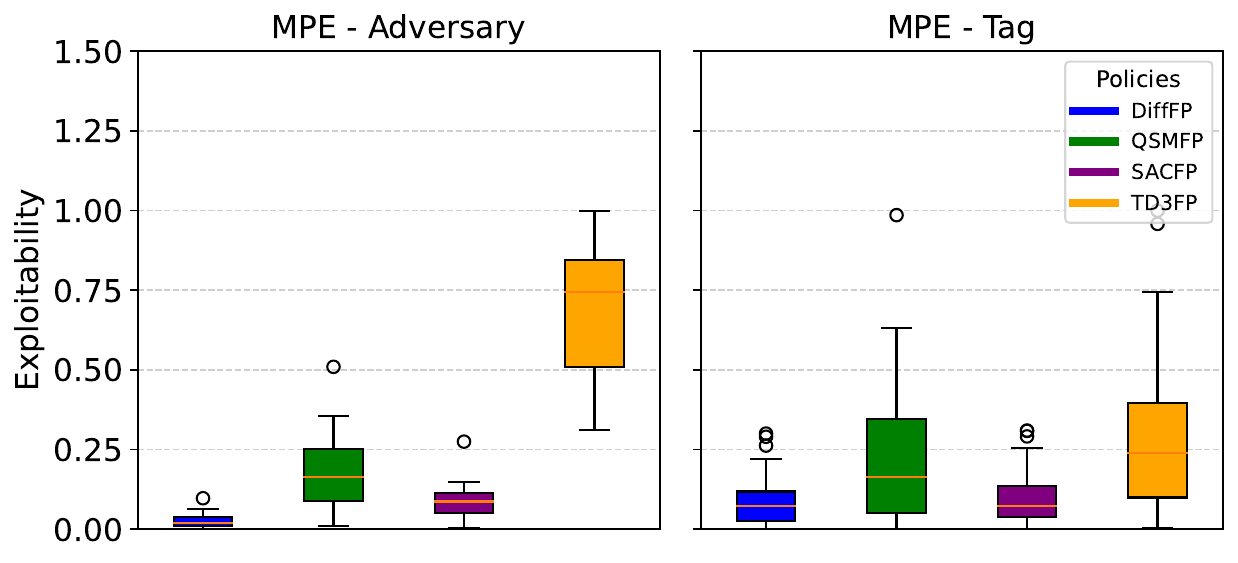}
    \caption{\textbf{Exploitability.} We report exploitability (see Def.~\ref{def:exp}) computed over 100 evaluation runs. Lower values indicate better robustness, with the proposed \model{} achieving the lowest exploitability.}
    \label{fig:mpe_adv}
\end{figure}

\subsection{\gls{mpe}}
\begin{figure*}[tb]
    \centering
    \includegraphics[width=\linewidth, clip]{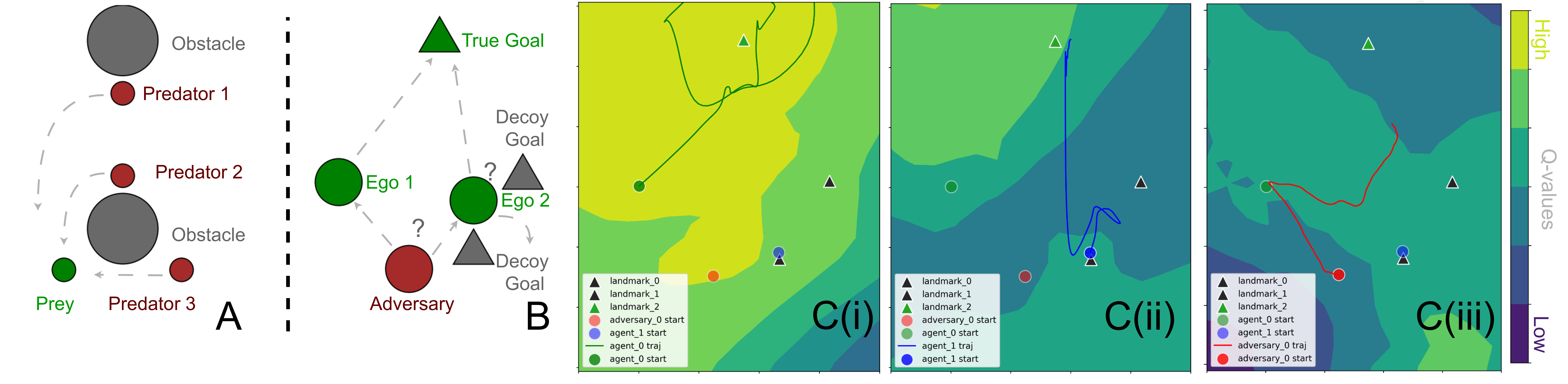}
    \caption{\textbf{A. MPE-Tag Env.}, \textbf{B. MPE-Adversary Env.} Dashed lines indicate possible agent trajectories. \textbf{C. Q-value Map.} At convergence to a near Nash equilibrium, ego agents exhibit coordinated behaviors with diverse roles e.g., one agent advances toward the goal while the other acts as a decoy to distract or mislead the adversaries.}
    \label{fig:mpe_value}
\vspace{-5mm}
\end{figure*}

To demonstrate the generalizability of our approach to broader multi-agent setups, we conduct experiments on the \gls{mpe} suite~\cite{lowe2017multi}. These tasks are particularly challenging due to their sparse reward structure and the existence of multi-modal solusions. We focus on two asymmetric and imperfect information scenarios:

\begin{itemize}
    \item \textbf{\gls{mpe}-Adversary:} Ego agents must collaboratively cover landmarks to mislead a single adversary from locating the hidden target landmark (Figure~\ref{fig:mpe_value}B).
    \item \textbf{\gls{mpe}-Tag:} Adversaries must coordinate to trap the ego agent, who must in turn evade capture (Figure~\ref{fig:mpe_value}A).
\end{itemize}
In both cases, agents operate under partial observability, receiving only local information about nearby entities. The environments are asymmetric in role and objective. We adopt centralized training with decentralized execution (CTDE) framework. This setup is particularly challenging~\cite{wang2023more}, as the ego agent is often required to adopt role-specific behaviors such as acting as a decoy which differ substantially from typical goal seeking strategies.
\subsubsection{\textbf{Results}}
We report the exploitability metric across 100 novel scenarios in Figure~\ref{fig:mpe_adv}. \model{} achieves the lowest exploitability with minimal variance, demonstrating strong generalization and stability. Qualitatively, only \model{} and SACFP variants converge to near-optimal performance, successfully learning cooperative behaviors. Figure~\ref{fig:mpe_value} shows the Q-value map, where one ego agent acts as a decoy to distract the adversary while the other reaches the goal. 

\noindent\paragraph{\textbf{Robustness to Unseen Opponents}} 
We evaluate robustness via cross-play matchups between all models (Tables~\ref{tab:dp_games} and~\ref{tab:algo_results}). In the \emph{MPE-Adversary} task, an ego win is recorded when the agent reaches the true goal without being caught; an adversary win occurs if it intercepts the goal-directed ego (catching a decoy still counts as failure), and a draw otherwise. In the \emph{MPE-Tag} task, the ego wins by successfully evading all adversaries, otherwise the adversaries win. \model{} consistently outperforms baselines when acting as the ego agent. When playing as the adversary, it improves baseline win rates and increases the number of draws, indicating strong generalization to unseen opponents.

\noindent\paragraph{\textbf{Population-Based Hardening}} In the \gls{mpe}-Adversary task, we evaluate against SACPBT (SAC with Population-Based Training)~\cite{gleave23}, which introduces two additional adversaries preventing ego from overfitting to a single strategy. This not only results in stronger adversaries, achieving a 90\% improvement over SACFP in the adversary role against \model{}, but also increases training diversity. However, we observe that \model{}, even without PBT, performs comparably against SACPBT, indicating its robustness to increasingly complex adversaries while avoiding the added computational cost.

\noindent\paragraph{\textbf{When Q-Value-Based Methods Excel}} In the \gls{mpe}-Tag setting, baseline performance is comparable to that of \model{}, particularly in the case of adversaries. This is expected, as we employ a team-centralized training setup. Given the structure of the Tag game with a single ego agent and three adversaries the setting effectively resembles a population-based hardening variant of 1v1 predator-prey game, which encourages the emergence of more robust behaviors. Additionally, Q-mode-seeking models tend to perform well as adversaries; with more samples, the estimation error in the Q-value function is reduced, improving overall adversarial performance (as seen in QSMFP).

\subsection{Discussion - Why Diffusion Policies ?} 
In imperfect information games optimal strategies are often stochastic and multi-modal. We highlight several key advantages of diffusion-based policies in this context:%
\begin{itemize}
    \item \textbf{Multi-modal Best Responses:} Optimal behavior often involves stochastic mixtures over diverse actions. Traditional unimodal policies (e.g., Gaussian actors) tend to collapse to a single dominant mode, whereas diffusion policies naturally support multi-modal action distributions, enabling richer and less predictable behaviors that are harder to exploit.
    
    \item \textbf{High-Entropy Stochasticity:} Diffusion models maintain high-entropy policies through sampling-based generation (\eqref{eq:reverse_sampling}). Unlike Gaussian policies, which inject noise post hoc and often sacrifice action quality, diffusion models produce diverse yet high-quality actions directly via denoising steps.    
\end{itemize}
\begin{table}[tb]
    \centering
    \renewcommand{\arraystretch}{1.2}
    \begin{tabular}{l l c c c}
        \hline
        \textbf{Ego} & \textbf{Adv} & \textbf{Ego Wins} & \textbf{Adv Wins} & \textbf{Draws} \\
        \hline
        \multicolumn{5}{c}{\textbf{Games where \model{} is Ego}} \\
        \hline
        \model{} & SACFP   & 79  & 10  & 11  \\
        \model{} & TD3FP   & 75  & 11  & 14  \\
        \model{} & QSMFP   & 75  & 12  & 13  \\
        \hline
        \multicolumn{5}{c}{\textbf{Games where \model{} is Adversary}} \\
        \hline
        SACFP  & \model{}  & 62 \textcolor{red}{(17$\downarrow$)} & 16 \textcolor{ForestGreen}{(6$\uparrow$)} & 22 \textcolor{ForestGreen}{(11$\uparrow$)} \\
        TD3FP  & \model{}  & 50 \textcolor{red}{(25$\downarrow$)} & 18 \textcolor{ForestGreen}{(7$\uparrow$)} & 32 \textcolor{ForestGreen}{(18$\uparrow$)} \\
        QSMFP  & \model{}  & 63 \textcolor{red}{(12$\downarrow$)} & 27 \textcolor{ForestGreen}{(15$\uparrow$)} & 10 \textcolor{red}{(3$\downarrow$)} \\
        \hline
    \end{tabular}
    \caption{\textbf{Robustness Against Unseen Opponents.} Head-to-head evaluation in the \textbf{\gls{mpe}-Adversary} environment. Colored numbers indicate relative gains \textcolor{ForestGreen}{({$\uparrow$})} or losses \textcolor{red}{({$\downarrow$})} when ego and adversary roles are swapped.}
    \label{tab:dp_games}
\end{table}

\begin{table}[tb]
    \centering
    \renewcommand{\arraystretch}{1.2}
    \begin{tabular}{l l c c}
        \hline
        \textbf{Ego} & \textbf{Adversary} & \textbf{Ego Wins} & \textbf{Adv Wins} \\
        \hline
        \model{}  & SACFP   & 81  & 19  \\
        \model{}  & TD3FP   & 89  & 11  \\
        \model{}  & QSMFP   & 64  & 36  \\
        \hline
        SACFP   & \model{}  & 75 & 25 \textcolor{ForestGreen}{(6$\uparrow$)} \\
        TD3FP   & \model{}  & 66 & 34 \textcolor{ForestGreen}{(23$\uparrow$)} \\
        QSMFP   & \model{}  & 68 & 32 \textcolor{red}{(4$\downarrow$)} \\
        \hline
        QSMFP   & SACFP   & 66  & 34  \\
        SACFP   & QSMFP   & 42  & 58 \textcolor{ForestGreen}{(24$\uparrow$)}  \\
        \hline
    \end{tabular}
    \caption{\textbf{Robustness Against Unseen Opponents.} Head-to-head evaluation in the \textbf{\gls{mpe}-Tag} environment. Colored numbers indicate relative gains \textcolor{ForestGreen}{({$\uparrow$})} or losses \textcolor{red}{({$\downarrow$})} when ego and adversary roles are swapped.}
    \label{tab:algo_results}
\end{table}

\section{Conclusion}
\label{sec:conclusion}

We study the problem of learning policies in dynamic and continuous games. We propose \model{}, a fictitious play framework that models the best response as a diffusion policy, capturing diverse and multi-modal action distributions. Extensive simulation studies demonstrate that \model{} achieves faster convergence, more stable training, and learns policies closer to equilibrium, while maintaining robust performance against diverse, unseen opponents. %

\noindent\textbf{Limitations.} While effective, our approach has some limitations:  
(1) Reward function is based on one-step utilities (e.g., track progress), which may overlook long-horizon strategic value.
(2) We do not explicitly control the agent’s racing style (e.g., aggressive vs. defensive), limiting behavioral diversity and safety. More structured control could improve interpretability and scalability in safety-critical systems.

\noindent\textbf{Future Work.} A promising direction is to incorporate trajectory-level optimization for safer and more strategic policy initialization via classical path planning. Finally, replacing the DDPM backbone with flow-matching techniques~\cite{lipman2022flow} may enable one-step denoising and significantly faster sampling, improving training and inference efficiency.

\small{
\bibliographystyle{ieee_fullname}
\bibliography{main}
}

\end{document}